\documentclass{article}
\usepackage{subcaption}
\usepackage{placeins}
\usepackage{float}
\usepackage{listings}

\usepackage{bbm}
\usepackage[preprint, nonanonymous]{neurips_2026}

\usepackage[utf8]{inputenc} 
\usepackage[T1]{fontenc}    
\usepackage{hyperref}       
\usepackage{url}            
\usepackage{booktabs}       
\usepackage{amsfonts}       
\usepackage{nicefrac}       
\usepackage{microtype}      
\usepackage{xcolor}         
\usepackage{graphicx}
\usepackage{adjustbox}
\usepackage{amsmath}
\usepackage{multirow}
\usepackage{bm}
\usepackage{siunitx}
\usepackage{colortbl}
\definecolor{auxrow}{gray}{0.93}   

\title{Masking Is Not Enough: Generative Restoration for Multimodal De-Identification in Medical AI
}

\author{%
  Shiva Shrestha \\
  Department of Computer Science \\
  Kennesaw State University \\
  Marietta, GA 30060, USA \\
  \texttt{sshres16@students.kennesaw.edu}
  \And
  Zongxing Xie \\
  Department of Computer Science \\
  Kennesaw State University \\
  Marietta, GA 30060, USA \\
  \texttt{zxie1@kennesaw.edu}
  \AND
  Chen Zhao \\
  Department of Computer Science \\
  Kennesaw State University \\
  Marietta, GA 30060, USA \\
  \texttt{czhao4@kennesaw.edu}
  \And
  Liran Ma \\
  Department of Computer Science and Software Engineering \\
  Miami University \\
  Oxford, OH 45056, USA \\
  \texttt{mal18@MiamiOH.edu}
  \AND
  Zhipeng Cai \\
  Department of Computer Science \\
  Georgia State University \\
  Atlanta, GA 30303, USA \\
  \texttt{zcai@gsu.edu}
  \And
  Honghui Xu \\
  Department of Information Technology \\
  Kennesaw State University \\
  Marietta, GA 30060, USA \\
  \texttt{hxu10@kennesaw.edu}
}

\begin{document}

\maketitle

\begin{abstract}
Medical image-text data can expose protected health information (PHI) through both visible image content as well as accompanying text, creating a barrier to privacy-preserving medical AI systems. This risk is especially prominent in multimodal systems, where images, questions, reports, and clinical context may enter training, evaluation, or inference pipelines. Existing medical vision-language benchmarks primarily emphasize task utility, while de-identification methods are often evaluated separately from downstream reasoning. We introduce ClinX, an end-to-end multimodal PHI sanitization framework for medical image-text data. ClinX detects visible identifiers with optical character recognition (OCR), constructs binary PHI masks, and applies ClinX-PRISM, a no-skip generative restoration module with privacy-oriented post-processing for burned-in identifier suppression. In parallel, text-side PHI is reduced through progressive de-identification levels: regex masking, context-aware masking, and rewrite-based sanitization. We evaluate ClinX in medical visual question answering (MedVQA), jointly measuring PHI leakage and downstream utility across image-side, text-side, and combined de-identification settings. Results show that OCR-only masking is not sufficient as a standalone solution, and restoration-based sanitization better preserves clinically relevant visual context while sharply reducing recoverable PHI.
\end{abstract}

\section{Introduction}
\label{sec:introduction}

Multimodal medical AI systems are increasingly being integrated into real-world clinical workflows, including diagnostics assistance, clinical documentations, patient triage, and as decision-supporting tools~\citep{poon2025aihealthcare}. A central challenge in deploying these medical AI systems is that they often depend on sensitive patient data containing the protected health information (PHI), such as patient names, dates of birth, medical record numbers, hospital identifiers, and other details that can reveal an individual's identity~\citep{poon2025aihealthcare, Faustini2026}. The risk is especially more prominent when hospitals export the medical imaging data for research or AI model development. Even after the Digital Imaging and Communications in Medicine (DICOM) metadata has been removed, the image pixels themselves may still contain burned-in annotations, including patient name, date of birth, medical record number, scan dates, and hospital identifiers rendered into the images directly~\citep{Macdonald2024-bx}. Similar leakage can also appear in scanned reports, free-text clinical notes, structured clinical queries, and image overlays. As a result, data that appears at the file or metadata level may still expose the patient's identity through visible text or associated language. This is not just a technical limitation, but also a regulatory concern. Under the Health Insurance Portability and Accountability Act (HIPAA) de-identification guidance, identifiers such as names, dates, and medical record numbers must be removed or properly managed before the data can be used for the research, or secondary analysis~\citep{hhs2012deidentification, ecfr164514}. Meeting this standard requires a systematic framework for identifying information across the full multimodal data pipeline.

Recent progress in medical AI has increasingly moved beyond the single-modality prediction and diagnosis towards the multimodal visual-language systems that connect the medical images with the questions, instructions, report understanding, and clinical contexts~\citep{chen2024surveymedicalvisionandlanguageapplications}. Medical vision-language models (VLMs) and multimodal medical large language models (LLMs) have shown growing promise for tasks such as image-grounded reasoning, report understanding, instruction following, and clinical question answering~\citep{li2024llavamed, Med-Flamingo, hulumed, medgemma, medvlthinker}. Within this broader direction, medical visual question answering (MedVQA) has become a focused benchmark for the evaluation of the model's ability to answer the natural-language questions grounded in medical imagery. Datasets such as VQA-RAD, PathVQA, and SLAKE have supported evaluation across the radiology, pathology, and knowledge-enhanced medical image understanding~\citep{VQA_RAD, he-etal-2021-towards, liu2021slakesemanticallylabeledknowledgeenhanceddataset}. However, most of this work has emphasized reasoning accuracy, answer correctness, and model capability, while privacy risk in the paired image-text input has received much less systematic research attention. In parallel, PHI protection has been studied through separate de-identification efforts, including DICOM metadata removal, OCR-based detection of burned-in text~\citep{Macdonald2024-bx, rempe2024medicaldeidentification}, as well as de-identification of free-text clinical notes~\citep{neamatullah2008automated, stubbs2015annotating}. These methods are essential to ensure the PHI does not leak in any way, shape, or form, but they are particularly evaluated as individual pre-processing or privacy-preserving steps rather than as a part of a downstream medical vision-language reasoning pipeline. As a result, it remains unclear how PHI removal from the image side and the question side jointly affects both privacy protection and model utility. This gap is central to MedVQA, where the image and the question may each contain clinically useful information while also serving as potential challenges for PHI leakage.

To address this gap, we introduce ClinX, an end-to-end multimodal PHI sanitization framework for the privacy preserving medical AI. ClinX is designed to operate before the medical image-text data are used for the model training, evaluation, or downstream inference, reducing the risk that patient identifiers are exposed to the learning and inference pipeline. Given PHI-contaminated multimodal inputs, ClinX produces sanitized image-text pairs through two coordinated branches. The image-side branch detects the visible identifiers with the optical character recognition (OCR), matches candidate text against PHI patterns, constructs the binary PHI mask, and then applies the ClinX-PRISM, a generative restoration and identifier suppression module that removes the visible PHI while preserving the surrounding clinical image structure. The text-side branch applies progressively stronger de-identification, moving from the regex-based masking, to context-aware masking, and finally a generative rewrite to sanitize the text from any PHI, and ensure the structure of the text is intact. Together, these two branches prevent the PHI-bearing visual and textual context from being passed directly into the multimodal medical AI models while preserving the clinical signal needed for learning and reasoning. In this paper, we instantiate and evaluate ClinX in the MedVQA setting, where sanitized image-question pairs are tested for both privacy protection and downstream answer utility.
The main contributions of this work are summarized as follows:
\begin{itemize}
    \item We introduce ClinX, an end-to-end multimodal PHI sanitization framework for medical image-text data before training, evaluation, or inference.
    \item ClinX-PRISM extends image-side de-identification beyond OCR-only masking through mask-conditioned no-skip generative restoration with privacy-oriented post-processing for burned-in identifier suppression.
    \item PHI leakage and answer utility are jointly evaluated in MedVQA across image-side, text-side, and combined multimodal de-identification settings.
\end{itemize}

We have organized the remainder of this paper as follows. Section~\ref{sec:related_work} reviews related work. Section~\ref{sec:method} presents ClinX framework. Section~\ref{sec:experiments} describes the experimental setup and reports the results from the experiment. Section~\ref{sec:conclusion} concludes the paper.

\section{Background and Related Work}
\label{sec:related_work}

\textbf{Medical AI} has increasingly shifted from the single-modality prediction towards multimodal vision-language systems that connect clinical images with questions, reports, instructions, and broader clinical contexts~\citep{moor2023foundation, hartsock2024vision, chen2024surveymedicalvisionandlanguageapplications}. Recent medical vision-language models (VLMs) have demonstrated strong performance on biomedical image conversation, few-shot medical VQA, radiology reasoning, and instruction following over medical images~\citep{li2024llavamed, Med-Flamingo, medgemma, hulumed, medvlthinker}. As image-text reasoning becomes central to medical AI, it also increases the number of channels through which protected health information can enter training, evaluation, or inference pipelines~\citep{chen-etal-2025-unveiling-privacy}. Compared with just the task utility and model downstream performance, this privacy risk has received far less systematic attention.

\textbf{Medical visual question answering (MedVQA)} offers a focused setting for studying image-text reasoning in clinical contexts. Datasets such as VQA-RAD~\citep{VQA_RAD}, PathVQA~\citep{he-etal-2021-towards}, SLAKE~\citep{liu2021slakesemanticallylabeledknowledgeenhanceddataset}, and PMC-VQA~\citep{Zhang2024}  have supported evaluation across radiology, pathology, and knowledge-enhanced medical image understanding. These datasets are primarily utility-centered, measuring answer accuracy, reasoning ability, and closed-ended question performance. They are also typically released in the pre-de-identified form and do not include the PHI annotations or controlled PHI-bearing image-text pairs, making them unsuitable for the systematic evaluation of the pipeline to detect, remove or preserve the utility after the PHI sanitization. As a result, there is no standard MedVQA benchmark for studying privacy leakage and answer utility together under the realistic multimodal PHI contamination.

\textbf{Clinical de-identification} has addressed PHI protection through a largely parallel line of work. Under the Health Insurance Portability and Accountability Act (HIPAA) Privacy Rule, de-identified health information must not identify an individual or provide a reasonable basis for doing so, with Safe Harbor and Expert Determination serving as the main regulatory pathways~\citep{ecfr164514, hhs2012deidentification}. In medical imaging, de-identification has commonly focused on removal of Digital Imaging and Communications in Medicine (DICOM) metadata, using optical character recognition (OCR) to detect the burned-in text, and mask visible identifiers, which could be the candidate for the PHI, before masking or redaction~\citep{Macdonald2024-bx, rutherford2021dicom}. Similarly, in clinical text approaches, de-identification has been studied through rule-based system, named-entity recognition, sequence labeling, and neural rewriting to suppress patient-identifying information from medical text records~\citep{neamatullah2008automated, stubbs2015annotating}. These methods are well established within their respective domains, but they are usually evaluated as standalone preprocessing steps. This separation is limiting for the multimodal medical AI, where an image may be sanitized while the paired text still contains the PHI, or the text may be sanitized while the image still exposes the burned-in identifiers.

\textbf{Masking mechanism} is one of practical ways for suppressing visible PHI, but it is not always sufficient for downstream image-text reasoning. OCR-based blackout masking can remove the surface-level identifiers, yet it may also disrupt the nearby visual structures that may be useful for the medical diagnosis and interpretation~\citep{Macdonald2024-bx, rutherford2021dicom}. Generative image inpainting offers a complementary direction by reconstructing masked regions from the learned visual context rather than simply masking or deleting them~\citep{kline2026redactionrestoration}. Foundational and widely used approaches include generative adversarial networks, context encoders, partial convolutions, DeepFill, and LaMa~\citep{Yu2018, Yu2019, liu2018image, suvorov2021resolutionrobustlargemaskinpainting, pathak2016context}. However, generic restoration is not the same as privacy-preserving sanitization; the recovered region must suppress identifiers while avoiding misleading visual artifacts and preserving downstream utility. This leaves multimodal PHI sanitization insufficiently studied, particularly when privacy leakage and task performance must be evaluated jointly across both image and text.

\begin{figure*}[t]
    \centering
    \includegraphics[width=\textwidth]{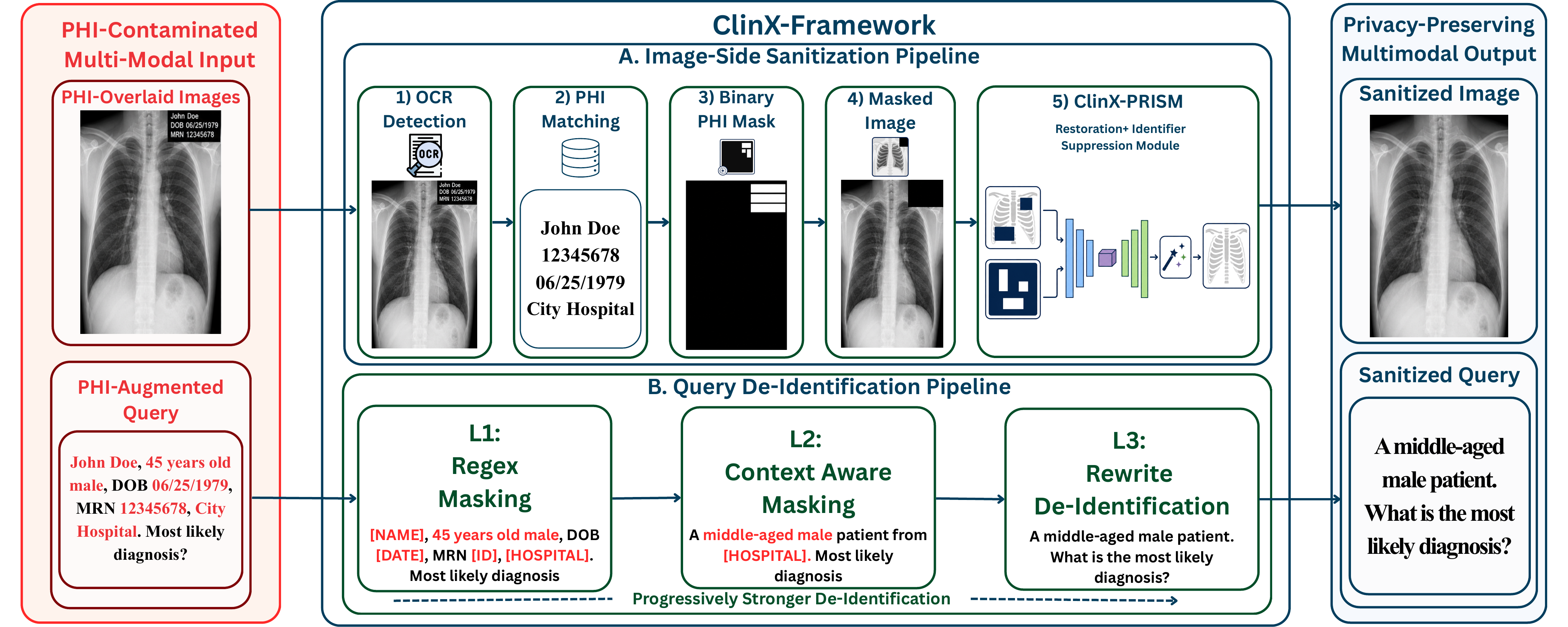}
    \caption{Overview of ClinX framework.}
    \label{fig:clinx_framework}
\end{figure*}

\section{ClinX}
\label{sec:method}
In this section, we introduce ClinX, our multimodal PHI sanitization pipeline for medical VQA. Specifically, our approach is designed to deal with instances where both images and questions may potentially contain PHI before being fed into the downstream model. In this section, we first introduce the threat model as well as the specific controlled environment of PHI-augmented image-question pairs. Next, we provide details on the image side of our pipeline, which leverages OCR localization and the use of our no-skip generative restoration module called ClinX-PRISM. Lastly, we explain our question de-identification procedure as well as our evaluation protocol.

\paragraph{Threat model}
We evaluate the potential for data misuse within a downstream data-use setting where medical image-question pairs containing PHI could become exposed to the model developers, VLM inference API, benchmark evaluators, or automatic extraction engines prior to de-identification. The adversary is not presumed to have access to the synthetic manifest utilized in our evaluation process, yet could exploit optical character recognition (OCR), string matching, or vision-language inference to extract any visible or textual identifiers from the published inputs. Any synthetic manifest identifier that could potentially be extracted from the sanitized medical image or question through the respective leakage evaluator is considered to be recovered. This threat model does not claim cryptographic privacy, rather it just targets practical PHI exposure through burned-in image text and question-side identifiers before downstream MedVQA inference.

\subsection{ClinX Overview}
\label{subsec:clinx_overview}
Our framework, ClinX, is designed for medical VQA settings in which protected health information (PHI) can appear in both the image and the accompanying questions. Given an image-question pair that is contaminated with PHI, ClinX generates a sanitized image and a sanitized question, which can then be fed into the downstream medical VLM. There are two goals we aim to achieve with our design: suppress recoverable PHI and preserve the clinical information essential for the correct answer generation.

Let us consider \(\mathcal{D}=\{(x_i,q_i,a_i)\}_{i=1}^{N}\) be a medical VQA dataset, where \(x_i\) is the image, \(q_i\) is the question, and \(a_i\) is the ground-truth answer. To study privacy and utility under controlled PHI exposure, we construct a PHI-augmented dataset as,
\begin{equation}
\widetilde{\mathcal{D}}
=
\{(\tilde{x}_i,\tilde{q}_i,a_i,m_i,z_i)\}_{i=1}^{N}.
\label{eq:phi_augmented_dataset}
\end{equation}
Here, \(\tilde{x}_i\) denotes the PHI-overlaid image, \(\tilde{q}_i\) denotes the PHI-augmented question, \(m_i\) is the binary mask for the image-side PHI region, and \(z_i\) is a shared PHI manifest containing the injected identity fields and placement metadata. The manifest ties two modalities together: it specifies which identifiers are inserted into the image, which identifiers are inserted into the question, where image-side PHI appears, and which ground-truth strings are used for leakage measurement. With the synthetic generation of all of the identifiers used in our pipeline, this scenario enables the privacy leakage evaluation without compromising any actual personal data of the patients.
In ClinX, there are two coupled sanitization branches as shown in the Fig.~\ref{fig:clinx_framework}. 

\subsection{Image-Side Sanitization Pipeline}
In the case of image branch, we first apply an OCR technique to identify text regions, followed by matching the candidate string to the shared manifest for constructing the PHI-only mask. The mask is then passed onto ClinX-PRISM, which then recreates the sensitive area while blurring out the burned-in identifiers.

\textbf{Image-Side PHI Localization.}
First of all, in the image-side branch, we localize the visible PHI in the PHI-overlaid image. As the injected identifiers generally appear as burned-in text in the medical images, OCR is used as a spatial text extraction operator. For each image denoted by \(\tilde{x}_i\), OCR returns detections \(O_i=\{(s_{ij},b_{ij},c_{ij})\}_{j=1}^{M_i}\), where \(s_{ij}\) is the recognized string, \(b_{ij}\) is its bounding region, and \(c_{ij}\) is the OCR confidence score. 
The shared manifest \(z_i=\{e_{ik}\}_{k=1}^{K_i}\) contains the synthetic identifiers inserted into the sample, including names, dates of birth, medical record numbers, and hospital identifiers. Thus, image-side localization is treated as manifest-conditioned PHI selection rather than generic text removal, which might remove all of the essential text from the medical images, including labels and descriptive texts.

For each of the OCR detection $O_i$, we compare the normalized recognized string against the manifest using 
\begin{equation}
    \alpha_{ij}=\max_k \mathrm{sim}(\nu(s_{ij}),\nu(e_{ik}))
\end{equation}
where \(\nu(\cdot)\) denotes the text normalization whereas \(\mathrm{sim}(\cdot,\cdot)\) measures exact or partial string overlap. A region is marked as PHI when \(\alpha_{ij}\geq \tau_{\mathrm{PHI}}\) and \(c_{ij}\geq \tau_{\mathrm{ocr}}\). We use this matching step to allow the minor OCR errors to be tolerated while avoiding unnecessary removal of all text-like structures from the image itself, which can indirectly destroy the meaningful information from the image will affect the downstream performance itself.
The binary mask \(m_i\) is then constructed by projecting the selected PHI bounding regions into image space and applying a small dilation around matched regions. The dilation accounts for OCR box imprecision and includes boundary pixels around the rendered identifier. As a direct redaction baseline, we define \(x_i^{\mathrm{ocr}}=R(\tilde{x}_i,m_i)=\tilde{x}_i\odot(1-m_i)+\kappa m_i\), where \(R(\cdot)\) denotes blackout redaction, \(\kappa\) is a constant blackout value, and \(\odot\) denotes element-wise multiplication. This baseline preserves all pixels outside the PHI mask and replaces only the localized identifier region.

This localization stage is intentionally conservative, but it remains bounded by OCR coverage. If an image-side PHI token is completely missed by OCR, it will not be included in the mask \(m_i\) and can thus remain visible in the sanitized image itself. Therefore, in ClinX, we do not assume perfect mask construction. Instead, residual leakages are measured after sanitization is applied using OCR-based privacy metrics. This limitation is important in practice and motivates reporting leakage explicitly rather than treating localization as a complete privacy guarantee.

\textbf{ClinX-PRISM: Privacy-Oriented Generative Restoration.}
Direct OCR redaction removes visible identifiers using the masking, thus preserving the privacy itself, but it also leaves a hard occlusion in the image. This is undesirable, particularly for medical images. It is so because the region surrounding a burned-in overlay may still carry useful visual context for downstream reasoning and diagnosis. We therefore introduce ClinX-PRISM, short for \textit{Privacy-preserving Restoration via Inpainting and Spatial Masking}, a privacy-oriented generative restoration module that replaces the localized PHI region while preserving the rest of the medical image unchanged. In ClinX, we therefore treat \textbf{generative restoration} not simply as
image reconstruction, but as a privacy-oriented \textbf{sanitization} process that removes
recoverable identifiers while preserving clinically relevant image context.
During the inference of the image-question pair, ClinX-PRISM receives only the PHI-overlaid image \(\tilde{x}_i\) and its PHI mask \(m_i\). The clean image \(x_i\) is used only during paired restoration training, where it provides supervision before synthetic PHI overlay. It is never available to the sanitizer at evaluation time and is not included as a main image condition itself, thus preventing any change of leakage of PHI.

We designed the ClinX-PRISM around a simple but important hypothesis, which is that the high-fidelity inpainting and privacy-preserving inpainting should not necessarily use the same information-flow pattern. In standard restoration, skip connections are often useful because they transfer high-resolution encoder features directly into the decoder, which we denote using \texttt{g-skip} notation. In PHI sanitization, however, the same pathway can be risky because encoder features are computed from the PHI-contaminated input; thus, the direct feature reuse may carry identifier-correlated strokes, edges, or local texture fragments into the restored region itself. We therefore treat skip removal as a privacy-oriented architectural constraint rather than only a reconstruction choice. This hypothesis is evaluated through the \texttt{g-skip} versus \texttt{ClinX-PRISM} comparison, where both models share the same training objective and differ primarily in generator information flow along with post-processing.
Given \(\tilde{x}_i\) and \(m_i\), ClinX-PRISM uses a no-skip generator (g-no) \(G_{\mathrm{P}}\) to predict a restored image \(\hat{y}_i=G_{\mathrm{P}}(\tilde{x}_i,m_i)\). The generated image is not used to replace the full input. Instead, the sanitized output is obtained through mask-restricted compositing:
\begin{equation}
\hat{x}_i
=
\hat{y}_i \odot m_i
+
\tilde{x}_i \odot (1-m_i).
\label{eq:prism_composition}
\end{equation}
This operation confines generation to the PHI-bearing region and copies all non-masked pixels directly from the original image to the newly generated image itself.
The generator follows a mask-conditioned encoder-decoder architecture. The encoder maps the input through channel widths \(3 \rightarrow 64 \rightarrow 128 \rightarrow 256 \rightarrow 512\), and the bottleneck contains three SPADE residual blocks at 512 channels. In ClinX, We use SPADE-style normalization because the PHI mask is a spatial conditioning signal rather than a global attribute itself~\citep{park2019semanticimagesynthesisspatiallyadaptive}. The mask which is denoted by $m_i$ specifies where the restoration should occur and where the original image should remain unchanged. Spatially adaptive normalization enables the generator to modulate feature statistics according to the mask at multiple resolutions, which is more suitable for localized PHI restoration than conditioning only through input concatenation. The decoder up-samples through \(512 \rightarrow 256 \rightarrow 128 \rightarrow 64\) using transpose-convolution stages followed by mask-conditioned residual refinement.

After compositing, we apply a mask-guided post-processing stage to the \texttt{g-no} output, which forms the final ClinX-PRISM output. 
For the post-processing, let us consider
\(c_i=\hat{y}_i\odot m_i+\tilde{x}_i\odot(1-m_i)\) denote the composited no-skip restoration. We first construct a soft refinement mask \(m_i^{f}\) by resizing the PHI mask to the image resolution and feathering its boundary with Gaussian smoothing. The post-processing stage is then written compactly as
\begin{equation}
\begin{array}{rcl}
m_i^{f} & = &
\mathrm{clip}\!\left(
\mathcal{B}_{\sigma_f}(\mathrm{Resize}(m_i)),0,1
\right),\\[2pt]
u_i & = &
c_i + m_i^{f}\odot \epsilon_i,\\[2pt]
v_i & = &
(1-m_i^{f})\odot u_i
+ m_i^{f}\odot (u_i\odot \gamma_i),\\[2pt]
w_i & = &
(1-m_i^{f})\odot v_i
+ m_i^{f}\odot \mathcal{B}_{\sigma_b}(v_i),\\[2pt]
x_i^{\mathrm{prism}} & = &
\mathcal{J}_{q}(w_i).
\end{array}
\label{eq:prism_postprocess}
\end{equation}
where,
$\mathcal{B}_{\sigma_f}$ denotes Gaussian feathering of the mask boundary,
$\epsilon_i \sim \mathcal{N}(0,\sigma_g^2)$ is shared-channel luminance grain,
$\gamma_i \sim \mathcal{N}(1,\sigma_c^2)$ is per-channel color jitter,
$\mathcal{B}_{\sigma_b}$ denotes lightweight Gaussian micro-blur, and
$\mathcal{J}_q$ denotes JPEG recompression at quality $q$. This stage introduces no learned parameters; it is only a passive refinement technique that processes the no-skip output by softening the restored boundary and reducing high-frequency remnants that OCR engines may interpret as text-like strokes, to generate the final ClinX-PRISM output. It is applied only to the ClinX-PRISM condition \texttt{g-no}, whereas \texttt{g-skip} uses masked compositing without this refinement, \texttt{ocr} uses direct blackout masking, and \texttt{raw} is left unchanged.

\textbf{Optimization and Image Conditions.}
In ClinX, We use an adversarial inpainting backbone as our evaluation requires the deterministic, mask-conditioned restoration over a large Cartesian grid of image and text privacy settings. Inpainting with a prompt conditioned diffusion model is another plausible choice since both approaches have been proven to be successful for the purpose of reconstruction, but it introduces additional prompt design, stochastic sampling, and reproducibility concerns. However, it is important to note that this paper does not intend to prove that one type of model is better than the other.
ClinX-PRISM module is trained as a paired adversarial inpainting model using tuples \((\tilde{x}_i,m_i,x_i)\), where \(x_i\) is the clean image before synthetic PHI overlay. We train the generator with a PatchGAN-style discriminator given by~\cite{demir2018patchbasedimageinpaintinggenerative} and combine adversarial supervision, masked reconstruction, identity preservation outside the mask, and total variation regularization:
\begin{equation}
\mathcal{L}_{\mathrm{PRISM}}
=
\lambda_{\mathrm{adv}}\mathcal{L}_{\mathrm{adv}}
+
\lambda_{\mathrm{mask}}\mathcal{L}_{\mathrm{mask}}
+
\lambda_{\mathrm{id}}\mathcal{L}_{\mathrm{id}}
+
\lambda_{\mathrm{tv}}\mathcal{L}_{\mathrm{tv}} .
\label{eq:prism_loss}
\end{equation}
Here, \(\mathcal{L}_{\mathrm{mask}}\) restores the PHI region, \(\mathcal{L}_{\mathrm{id}}\) discourages changes outside the mask, and \(\mathcal{L}_{\mathrm{tv}}\) reduces local artifacts near the restored boundary. The loss weights and restoration hyperparameters are fixed across restoration variants and reported in the appendix.
For image-side sanitization, we evaluate four conditions:
$
\mathcal{C}_{\mathrm{img}}
=
\{\texttt{raw},~\texttt{ocr},~\texttt{g-skip},~\texttt{ClinX-PRISM}\}.
$
Here, \texttt{raw} denotes the PHI-overlaid image, \texttt{ocr} represents the OCR-guided blackout masking, \texttt{g-skip} is the skip-connected restoration ablation, and \texttt{ClinX-PRISM} denotes the proposed ClinX-PRISM pathway with no-skip restoration and mask-guided post-processing.


\subsection{Progressive Query De-Identification Pipeline}
\label{subsec:text_deid}
In the case of question text data, ClinX performs text-level de-identification using increasingly powerful de-identification steps, including pattern-based and context-aware masking, as well as rewrites. Finally, both the sanitized images and texts are combined and validated through the downstream MedVQA.
In ClinX framework, we use the question-side branch to sanitize the PHI through four privacy levels, \(L_{\mathrm{text}}=\{L_0,L_1,L_2,L_3\}\). At \(L_0\), the PHI-augmented question is retained, which is used to isolates the effect of image-side sanitization. At \(L_1\), explicit surface-form identifiers are masked with regex-based rules, including names, dates, record numbers, phone numbers, email addresses, and hospital identifiers. Following \(L1\) sanitization, at \(L_2\), context-aware masking is applied to suppress residual PHI that may appear near anchors such as ``Patient,'' ``Name,'' ``DOB,'' ``MRN,'' ``ID,'' or ``Hospital.'' This stage expands sanitization from direct string matching to context-sensitive PHI removal.
Finally, for \(L_3\), we generate rewrite-based de-identification manifests using an offline large language model (LLM). We instruct the model to remove all of the remaining PHI like names, dates, record numbers, hospital identifiers, locations, and other identity-bearing phrases if any, while preserving the medical question being asked. In practice, we preserve the clinical finding, anatomy, imaging modality, disease cue, and question intent whenever they are present, and remove demographic or institutional details that are not needed to answer the VQA itself. Rewrites are generated before evaluation and are not particularly adapted to downstream VLM predictions.
Formally, starting from \(\tilde{q}_i^{(0)}=\tilde{q}_i\), the text transformations are \(\tilde{q}_i^{(1)}=T_1(\tilde{q}_i^{(0)})\), \(\tilde{q}_i^{(2)}=T_2(\tilde{q}_i^{(1)})\), and \(\tilde{q}_i^{(3)}=T_3(\tilde{q}_i^{(2)})\), where \(T_1\), \(T_2\), and \(T_3\) denote regex masking, context-aware masking, and rewrite-based de-identification, respectively.

\subsection{Joint Multimodal Evaluation Design}
\label{subsec:joint_design}

In ClinX, both image-side and question-side sanitization are tested simultaneously. For every image-question pair in ClinX, the Cartesian product \(\mathcal{C}_{\mathrm{img}}\times L_{\mathrm{text}}\) is used, which connects every image condition to every text privacy level. This approach enables us to study the effect of image sanitization in isolation from combined multimodal sanitization, since all \(L_{1\leq x\leq 3}\) de-identify the question and only \(L_0\) preserves PHI-augmented questions to isolate image-side sanitization.
Since the manifest \(z_i\) governs both the image as well as the question modality, this is not merely a post hoc grid of pre-processing choices. It measures how image restoration and question de-identification interact under a shared privacy process. A separate text-only ablation study is also performed where the image is left unchanged, but only the question is manipulated across different privacy levels. This way, we can better understand the effect of question-side sanitization independent of image restoration, and help us better interpret the privacy-utility tradeoff in MedVQA.
Additional details on synthetic PHI injection, shared manifest construction,
OCR matching, and mask construction are provided in
Appendix~\ref{app:benchmark_details}.

\section{Experiment and Evaluation}
\label{sec:experiments}
In this section, we present the experimental setup and evaluation results of ClinX. We first describe the datasets, privacy settings, medical VLMs, training details, and evaluation metrics used in our experiments. We then report the performance of ClinX, analyze the effectiveness of the proposed text de-identification mechanism, and provide a detailed privacy protection analysis.

\subsection{Experimental Setup}
\label{subsec:exp_setup}

\textbf{Datasets.} We evaluated ClinX on PHI-augmented versions of three popular MedicalVQA datasets PathVQA, VQA-RAD, and SLAKE. Augmentation was done by injecting the synthetic PHI into both the image and question modalities using the shared manifest described in Section~\ref{sec:method}. The test split contained 6,719 PathVQA examples, 451 VQA-RAD examples, and 1,061 SLAKE examples. The training split was used only to train the restoration models, the validation split was used for checkpoint selection, and the test split was reserved for MedVQA and privacy evaluation, to ensure there is no data leakage.

\textbf{Privacy Conditions.} On the image side, we evaluated four primary conditions:
\(
\mathcal{C}_{\mathrm{img}}
=
\{\texttt{raw},\texttt{ocr},\texttt{g-skip},\texttt{PRISM}\}.
\)
where, \texttt{raw} is the PHI-overlaid image, \texttt{ocr} is just OCR-guided blackout masking, \texttt{g-skip} is the skip-connected restoration baseline, and \texttt{PRISM} denotes ClinX-PRISM, the proposed no-skip restoration pathway with mask-guided post-processing. For OCR leakage analysis only, we also report \texttt{PRISM-flat}, a no-postprocessing diagnostic export of the same no-skip generator, which is excluded from utility comparisons. On the text side, we evaluated \(L_{\mathrm{text}}=\{L0,L1,L2,L3\}\), corresponding to the PHI-augmented question, regex masking, context-aware masking, and rewrite-based de-identification. Finally, we evaluated the full Cartesian product \(\mathcal{C}_{\mathrm{img}}\times L_{\mathrm{text}}\).

\textbf{Base Medical VLMs.} We evaluated four medical VLMs: LLaVA-Med-7B~\citep{li2024llavamed}, MedGemma-1.5-4B~\citep{medgemma}, MedVLThinker-3B~\citep{medvlthinker}, and HuluMed-4B~\citep{hulumed}.

\textbf{Training Configurations.} Both restoration variants were trained with the same PatchGAN-style adversarial objective, using 10 epochs, batch size 6, learning rate \(10^{-4}\), input resolution \(224\times224\), and BF16 mixed precision on 2\(\times\) NVIDIA A100 80GB GPUs. 

\textbf{Performance Metrics.} Utility was measured using exact match (EM), token-level F1 (F1), yes/no exact match (Yes/no EM), and LLM-as-a-Judge scores with Llama 3.1-8B~\citep{grattafiori2024llama3herdmodels} and Mistral-7B-Instruct-v0.3~\citep{jiang2023mistral7b}. Privacy leakage was measured using OCR-based PHI token recall and character overlap between injected PHI strings and OCR-recovered text.
Implementation details for ClinX-PRISM and the LLM-as-a-Judge prompting and
scoring protocol are provided in Appendix~\ref{app:impl}.

\begin{table*}[t]
\centering
\setlength{\tabcolsep}{4.5pt}
\renewcommand{\arraystretch}{1.08}

\caption{%
  Privacy--utility tradeoff across text-privacy levels (\(L0\)--\(L3\)) and
  image conditions.
  \textbf{Bold} marks the best value per metric and level among the four
  primary conditions (\texttt{raw}, \texttt{ocr}, \texttt{g-skip},
  \texttt{PRISM}).
}
\label{tab:main_results_pair}

\vspace{4pt}

\begin{minipage}[t]{0.605\textwidth}
  \centering\footnotesize
  \setlength{\tabcolsep}{3.5pt}
  \textbf{(a)~Utility} \quad {\footnotesize($\uparrow$ higher is better)}
  \vspace{4pt}

  \begin{tabular}{%
    l
    l
    S[table-format=1.2]
    S[table-format=2.2]
    S[table-format=2.2]
    S[table-format=2.2]
    S[table-format=2.2]
  }
  \toprule
  {Lvl} & {Cond.}
        & {EM}
        & {F1}
        & {Y/N\,EM}
        & {Llama-J}
        & {Mistral-J} \\
  \midrule

  \multirow{4}{*}{$L0$}
    & raw    & 6.72 &  9.67 & 61.85 & \bfseries 60.50 & \bfseries 45.34 \\
    & ocr    & 6.67 &  9.61 & 61.67 & 59.23 & 44.66 \\
    & g-skip & 8.19 & 11.16 & 63.83 & 59.42 & 45.12 \\
    & PRISM  & \bfseries 8.22 & \bfseries 11.18 & \bfseries 63.98 & 59.30 & 44.78 \\
  \midrule

  \multirow{4}{*}{$L1$}
    & raw    & 6.78 &  9.68 & 61.73 & \bfseries 59.09 & 43.35 \\
    & ocr    & 6.74 &  9.63 & 61.42 & 57.81 & 42.73 \\
    & g-skip & \bfseries 8.01 & \bfseries 11.01 & \bfseries 64.37 & 58.45 & \bfseries 43.82 \\
    & PRISM  & 7.97 & 10.95 & 64.21 & 58.09 & 43.40 \\
  \midrule

  \multirow{4}{*}{$L2$}
    & raw    & 6.61 &  9.66 & 61.82 & 61.51 & 43.49 \\
    & ocr    & 6.54 &  9.58 & 61.37 & 60.19 & 42.97 \\
    & g-skip & 8.53 & 11.53 & 63.70 & 61.49 & \bfseries 43.75 \\
    & PRISM  & \bfseries 8.57 & \bfseries 11.55 & \bfseries 63.82 & \bfseries 61.68 & 43.55 \\
  \midrule

  \multirow{4}{*}{$L3$}
    & raw    & 3.91 &  7.47 & 65.34 & 62.49 & 52.63 \\
    & ocr    & 3.88 &  7.48 & 65.17 & 62.06 & 52.58 \\
    & g-skip & \bfseries 7.15 & \bfseries 10.46 & \bfseries 67.01 & \bfseries 62.76 & \bfseries 53.91 \\
    & PRISM  & 7.11 & 10.38 & 66.78 & 62.61 & 53.56 \\
  \bottomrule
  \end{tabular}
\end{minipage}%
\hspace{0.012\textwidth}%
\begin{minipage}[t]{0.358\textwidth}
  \centering\footnotesize
  \setlength{\tabcolsep}{3pt}
  \textbf{(b)~Privacy} \quad {\footnotesize($\downarrow$ lower is better)}
  \vspace{4pt}

\resizebox{\linewidth}{!}{
  \begin{tabular}{llrr}
  \toprule
  & & \multicolumn{1}{c}{Tok.\ Rec.} & \multicolumn{1}{c}{Char Ovl.} \\
  \multicolumn{1}{l}{Eng.} & Variant & \multicolumn{1}{c}{(\%)} & \multicolumn{1}{c}{(\%)} \\
  \midrule

  \multirow{4}{*}{Easy}
    & raw              & 37.99               & 34.39               \\
    & ocr              & 0.054               & 0.029               \\
    & g-skip           & $\bm{3.2\times10^{-4}}$ & $\bm{1.3\times10^{-4}}$ \\
    & PRISM            & $1.6\times10^{-3}$  & $6.8\times10^{-4}$  \\
  \cmidrule(lr){2-4}
  \rowcolor{auxrow}
    &
    \textit{PRISM-flat} & \textit{$2.1\times10^{-4}$} & \textit{$1.1\times10^{-4}$} \\
  \midrule

  \multirow{4}{*}{Rapid}
    & raw              & 40.92               & 36.04               \\
    & ocr              & 0.029               & 0.018               \\
    & g-skip           & $\bm{1.08\times10^{-2}}$ & $\bm{5.3\times10^{-3}}$ \\
    & PRISM            & $1.45\times10^{-2}$ & $6.8\times10^{-3}$  \\
  \cmidrule(lr){2-4}
  \rowcolor{auxrow}
    &
    \textit{PRISM-flat} & \textit{$1.4\times10^{-3}$} & \textit{$6.9\times10^{-4}$} \\
  \bottomrule
  \end{tabular}
}
\end{minipage}

\vspace{5pt}
\parbox{0.96\textwidth}{%
  \footnotesize
  \textit{Abbreviations:}~%
  ocr\,=\,OCR-guided blackout masking;~%
  g-skip\,=\,skip-connected restoration baseline;~%
  PRISM\,=\,ClinX-PRISM, the proposed no-skip restoration pathway with mask-guided post-processing;~%
  PRISM-flat\,=\,auxiliary no-postprocessing OCR retest using the same no-skip generator.
}
\end{table*}

\subsection{Overall Privacy--Utility Tradeoff}
\label{subsec:overall_tradeoff}
In Table~\ref{tab:main_results_pair}, we observe an evident privacy-utility tradeoff across the evaluated sanitization settings. First, OCR-guided blackout masking drastically decreased the amount of recoverable PHI compared to the \texttt{raw} baseline conditions but performed comparably to and in some cases just under the performance of the \texttt{raw} baselines when considering EM and F1 scores. This result highlights the inadequacy of directly applying blackout masking to maintain MedVQA utility downstream.
On the other hand, the restore conditions allowed us to keep high utility levels while keeping OCR leakage close to zero.
When comparing among the three levels of text sanitization \(L0-L3\), \texttt{g-skip} and \texttt{PRISM} were able to outperform both \texttt{raw} and \texttt{ocr} when it comes to lexical metrics, especially under stricter text-side privacy settings. These results support the main premise of ClinX, that effective multimodal PHI sanitization should suppress identifiers while preserving clinically useful image context. Dataset-specific and model-wise results are provided in Appendix~\ref{app:additional_results}.

\subsection{Effect of Image-Side Sanitization}
\label{subsec:image_sanitization_effect}

The image-side experiment captures the difference between redaction and restoration. The \texttt{ocr} condition removed the detected PHI region through hard blackout masking, but this do not result in better downstream MedVQA reasoning. On the other hand, both the \texttt{g-skip} and \texttt{PRISM} conditions restored visual cues and perform consistently better in terms of EM and F1 compared to \texttt{raw} and \texttt{ocr} on \(L0-L3\) experiments. This is particularly true for conditions with stricter text-side privacy levels, as the need to retain image features becomes more crucial. Although \texttt{PRISM} does not uniformly outperform \texttt{g-skip}, it remains competitive while enforcing the no-skip, privacy-oriented restoration pathway used in ClinX.

\subsection{Effect of Progressive Query De-Identification}
\label{subsec:text_progression}
Utility stayed relatively constant between \(L0\) and \(L2\), indicating that both regex- and context-based identification removal does not change the clinical nature of the questions itself. Utility is most affected at \(L3\), when de-identification through rewrites makes changes to the question's structure itself. Even in this harder configuration, restoration-based image conditions still outperform \texttt{raw} and \texttt{ocr} conditions. This result indicates that, as sanitization becomes more restrictive on the question side, the importance of maintaining clinical information on the image side increases.

\subsection{OCR-Based Privacy Protection}
\label{subsec:ocr_leakage}
The results shown in Table~\ref{tab:main_results_pair}(b) confirmed that \texttt{raw} images without any protection contained considerable amounts of PHI that were recoverable via OCR engines, with token recall rates higher than 37\%. Both methods of protection drastically decreased token recall rates, demonstrating that both masking and restoration-based sanitization effectively reduced OCR recoverability. Within restoration-based sanitization, the \texttt{g-skip} variant provided the least amount of leakages among all of the options, while the post-processed version of \texttt{PRISM} still yielded slightly worse results. The supplementary analysis of \texttt{PRISM-flat} suggested that this phenomenon was related to the processing phase rather than the no-skip backbone.


\section{Discussion}
\label{sec:discussion}
Our proposed framework, ClinX treats the PHI sanitization in MedVQA as a multimodal privacy-utility problem, rather than simply as image redaction and anonymization. This importance of this distinction becomes evident from the findings, where OCR-blackout masking effectively removes visible identifiers from the image, yet it fails to restore the local visual context that can be useful for reasoning in downstream tasks. In contrast, restoration-based sanitization performed consistently better by offering a better balance by reducing OCR-recoverable PHI, while also preserving more of the image evidence needded for the downstream reasoning.
Additionally, a comparison of \texttt{g-skip} and \texttt{PRISM} highlights the importance of the no-skip architecture. The no-skip pathway is not necessarily a universal improvement in terms of the increase in accuracy when compared to the skip-connected baseline. However, it offers a comparable performance while leveraging a privacy-driven restoration pathway that prevents direct feature reuse from the PHI-contaminated input. Nevertheless, the ClinX framework offers no complete guarantees in terms of privacy, since leakage detection was limited on OCR-based recovery and synthetic PHI. Additional privacy audits and studies involving actual PHI patterns remains to be conducted in the future. Qualitative examples and additional limitations of the privacy evaluation are provided in Appendix~\ref{app:qual_limit}.

\section{Conclusion}
\label{sec:conclusion}
In this paper, we presented ClinX, a multimodal PHI sanitization framework for MedVQA that jointly addresses both image-side and text-side identifiers. Across datasets, privacy settings, and evaluated medical vision-language models, restoration-based sanitization substantially reduced the OCR-recoverable protected health information than blackout-masking methods, while consistently retaining higher utility. This paper demonstrates that privacy-preserving MedVQA should be evaluated as a joint multimodal problem, and not merely the redaction of images alone. Future work will extend ClinX toward stronger privacy audits, non-text PHI, and broader clinical multimodal settings.



\bibliographystyle{unsrtnat}
\bibliography{references}

\appendix

\section*{Appendix Overview}
\label{sec:appendix_overview}

This appendix section provides additional experimental detail and supporting analyses for ClinX framework. In Section~\ref{app:additional_results}, we report expanded utility and privacy results, including full condition-wise breakdowns, the auxiliary flattened \texttt{PRISM} leakage retest, and additional model-wise tables. Section~\ref{app:benchmark_details} documents the process of construction of the multimodal PHI benchmark, including synthetic PHI injection, OCR-based PHI matching, and mask construction. Then, Section~\ref{app:impl} gives implementation details for ClinX-PRISM and the LLM-as-a-Judge prompting and scoring protocol. Finally, in Section~\ref{app:qual_limit}, we provide qualitative examples, where available, and summarizes important limitations of the privacy evaluation.

\section{Additional Results}
\label{app:additional_results}

\subsection{Dataset-Specific Utility Breakdown}
\label{app:dataset_utility}
In Tables~\ref{tab:dataset_utility_all}~(a)-(c), we report dataset-specific utility breakdowns across text privacy levels and image conditions. These appendix tables complement the aggregate analysis in the main text by dis-aggregating results for 3 separate dataset PathVQA, VQA-RAD, and SLAKE, and help show whether the same privacy-utility pattern holds within each benchmark individually.
\begin{table*}[t]
\centering
\caption{Dataset-specific utility breakdowns across text privacy levels and image conditions. Higher is better.}
\label{tab:dataset_utility_all}

\begin{minipage}[t]{0.48\textwidth}
\centering
\footnotesize
\setlength{\tabcolsep}{4pt}
\subcaption{PathVQA}
\label{tab:pathvqa_levels}
\begin{tabular}{llccccc}
\toprule
Lvl. & Cond. & EM & F1 & Y/N EM & Llama-J & Mistral-J \\
\midrule
L0 & raw    & 7.16 & 9.59 & 62.24 & 60.02 & 43.10 \\
   & ocr    & 7.12 & 9.53 & 61.90 & 58.70 & 42.31 \\
   & g-skip & 8.67 & 11.12 & 64.09 & 58.40 & 42.53 \\
   & PRISM  & 8.70 & 11.16 & 64.27 & 58.37 & 42.20 \\
\midrule
L1 & raw    & 7.18 & 9.53 & 61.94 & 58.36 & 40.90 \\
   & ocr    & 7.20 & 9.53 & 61.51 & 56.97 & 40.19 \\
   & g-skip & 8.50 & 10.93 & 64.57 & 57.31 & 41.11 \\
   & PRISM  & 8.45 & 10.88 & 64.38 & 56.92 & 40.57 \\
\midrule
L2 & raw    & 7.44 & 9.87 & 62.07 & 61.12 & 41.14 \\
   & ocr    & 7.37 & 9.78 & 61.69 & 59.69 & 40.57 \\
   & g-skip & 9.26 & 11.66 & 64.38 & 60.86 & 41.08 \\
   & PRISM  & 9.28 & 11.68 & 64.34 & 61.06 & 40.89 \\
\midrule
L3 & raw    & 4.40 & 7.55 & 66.17 & 62.78 & 52.08 \\
   & ocr    & 4.33 & 7.50 & 65.91 & 62.38 & 52.04 \\
   & g-skip & 7.69 & 10.56 & 67.58 & 62.39 & 52.96 \\
   & PRISM  & 7.66 & 10.50 & 67.47 & 62.24 & 52.56 \\
\bottomrule
\end{tabular}
\end{minipage}
\hfill
\begin{minipage}[t]{0.48\textwidth}
\centering
\footnotesize
\setlength{\tabcolsep}{4pt}
\subcaption{VQA-RAD}
\label{tab:vqarad_levels}
\begin{tabular}{llccccc}
\toprule
Lvl. & Cond. & EM & F1 & Y/N EM & Llama-J & Mistral-J \\
\midrule
L0 & raw    & 7.10 & 12.50 & 61.70 & 61.92 & 53.99 \\
   & ocr    & 7.37 & 12.66 & 62.27 & 60.53 & 53.60 \\
   & g-skip & 7.88 & 13.10 & 61.43 & 62.31 & 53.16 \\
   & PRISM  & 7.82 & 13.09 & 62.30 & 62.64 & 53.55 \\
\midrule
L1 & raw    & 7.48 & 13.00 & 63.44 & 62.69 & 53.71 \\
   & ocr    & 7.32 & 12.66 & 63.40 & 60.03 & 52.00 \\
   & g-skip & 7.54 & 13.11 & 64.18 & 62.09 & 52.89 \\
   & PRISM  & 7.60 & 13.03 & 64.18 & 61.26 & 53.50 \\
\midrule
L2 & raw    & 4.60 & 10.94 & 64.23 & 62.14 & 52.99 \\
   & ocr    & 4.71 & 10.93 & 62.74 & 61.09 & 51.83 \\
   & g-skip & 7.54 & 13.18 & 60.40 & 61.98 & 52.28 \\
   & PRISM  & 7.71 & 13.32 & 60.80 & 63.09 & 52.50 \\
\midrule
L3 & raw    & 0.55 & 6.98 & 61.78 & 61.75 & 54.10 \\
   & ocr    & 0.55 & 7.03 & 61.77 & 60.92 & 53.27 \\
   & g-skip & 3.44 & 9.31 & 63.09 & 64.14 & 55.66 \\
   & PRISM  & 3.39 & 9.30 & 63.19 & 64.86 & 56.27 \\
\bottomrule
\end{tabular}
\end{minipage}

\vspace{0.8em}

\begin{minipage}[t]{0.60\textwidth}
\centering
\footnotesize
\setlength{\tabcolsep}{4pt}
\subcaption{SLAKE}
\label{tab:slake_levels}
\begin{tabular}{llccccc}
\toprule
Lvl. & Cond. & EM & F1 & Y/N EM & Llama-J & Mistral-J \\
\midrule
L0 & raw    & 3.79 & 8.97 & 59.09 & 62.96 & 55.89 \\
   & ocr    & 3.49 & 8.78 & 59.64 & 61.99 & 55.73 \\
   & g-skip & 5.28 & 10.54 & 63.55 & 64.63 & 58.10 \\
   & PRISM  & 5.32 & 10.48 & 63.02 & 63.78 & 57.42 \\
\midrule
L1 & raw    & 3.96 & 9.23 & 59.01 & 62.23 & 54.50 \\
   & ocr    & 3.56 & 8.94 & 59.52 & 62.21 & 54.92 \\
   & g-skip & 5.13 & 10.60 & 63.07 & 64.09 & 57.11 \\
   & PRISM  & 5.06 & 10.48 & 63.03 & 64.13 & 57.07 \\
\midrule
L2 & raw    & 2.21 & 7.78 & 58.50 & 63.71 & 54.31 \\
   & ocr    & 2.10 & 7.78 & 58.26 & 62.98 & 54.38 \\
   & g-skip & 4.36 & 9.98 & 61.04 & 65.29 & 57.04 \\
   & PRISM  & 4.43 & 9.97 & 62.07 & 64.98 & 56.57 \\
\midrule
L3 & raw    & 2.24 & 7.21 & 61.06 & 60.98 & 55.47 \\
   & ocr    & 2.47 & 7.57 & 61.53 & 60.53 & 55.73 \\
   & g-skip & 5.35 & 10.28 & 65.03 & 64.46 & 59.21 \\
   & PRISM  & 5.20 & 10.04 & 63.66 & 63.97 & 58.72 \\
\bottomrule
\end{tabular}
\end{minipage}

\vspace{4pt}
\parbox{0.92\textwidth}{%
  \footnotesize
  \textit{Abbreviations:}~%
  ocr\,=\,OCR-guided blackout masking;~%
  g-skip\,=\,skip-connected restoration baseline;~%
  PRISM\,=\,ClinX-PRISM, the proposed no-skip restoration pathway with mask-guided post-processing.
}

\end{table*}
\subsection{Auxiliary \texttt{PRISM-flat} No-Postprocessing Leakage Retest}
\label{app:prism_flat}

We have provided the main results in Table~\ref{tab:main_results_pair}(b), which includes an auxiliary no-postprocessing variant of ClinX-PRISM, denoted \texttt{PRISM-flat}. This variant uses the same no-skip restoration backbone, as that of the primary \texttt{PRISM} condition, but doesn't include the final mask-guided post-processing stage. We introduced this diagnostic control to separate two factors that are coupled in the primary ClinX-PRISM output: the no-skip restoration architecture and then, the final rendered post-processing stage.

In the primary comparison, \texttt{PRISM} showed slightly higher residual OCR leakage than \texttt{g-skip}. We do not interpret this result as direct evidence that the no-skip architecture is intrinsically less private. A second explanation is that the post-processing stage can introduce faint boundary, grain, or even compression like artifacts that OCR engines occasionally parse as text-like structure. From the \texttt{PRISM-flat} retest, we can probe this distinction by retaining the same no-skip generator while removing just the final post-processing stage at evaluation time.

Under both OCR engines, which were EasyOCR and RapidOCR, \texttt{PRISM-flat} produced comparitively lower residual leakage than the primary \texttt{PRISM} output. This supports our interpretation that the modest OCR disadvantage of the \texttt{PRISM} architecture is more plausibly associated with the final rendered output than with the no-skip restoration backbone alone. On the other hand, \texttt{PRISM-flat} is not treated as a primary condition because it was evaluated only for OCR leakage and not for downstream MedVQA utility. Its role was strictly limited to just the diagnostic, as it helps to explain the leakage behavior of the post-processed ClinX-PRISM output without changing the primary evaluation set overall.

\begin{table*}[t]
\centering
\footnotesize
\setlength{\tabcolsep}{3.5pt}
\caption{Expanded model-wise utility results across text privacy levels. Each subtable corresponds to one question-side privacy level. Higher is better for all reported metrics. \texttt{PRISM} denotes ClinX-PRISM, the proposed no-skip restoration pathway with mask-guided post-processing.}
\label{tab:appendix_modelwise_levels}

\begin{subtable}[t]{0.49\textwidth}
\centering
\caption{\(L0\)}
\resizebox{\linewidth}{!}{%
\begin{tabular}{llcccc}
\toprule
Model & Cond. & EM & F1 & Llama-J & Mistral-J \\
\midrule
\multirow{4}{*}{HuluMed-4B}
& raw    & 26.90 & 29.42 & 51.85 & 42.92 \\
& ocr    & 26.67 & 29.24 & 51.57 & 42.19 \\
& g-skip & 28.75 & 31.57 & 56.82 & 47.62 \\
& PRISM  & 28.87 & 31.68 & 56.87 & 47.58 \\
\midrule
\multirow{4}{*}{LLaVA-Med-7B}
& raw    & 0.00 & 6.20 & 54.95 & 41.47 \\
& ocr    & 0.00 & 6.20 & 54.83 & 41.61 \\
& g-skip & 0.00 & 10.21 & 58.76 & 45.21 \\
& PRISM  & 0.00 & 10.28 & 58.87 & 45.33 \\
\midrule
\multirow{4}{*}{MedGemma-1.5-4B}
& raw    & 0.00 & 2.01 & 70.05 & 50.41 \\
& ocr    & 0.00 & 2.03 & 68.36 & 49.71 \\
& g-skip & 0.00 & 5.95 & 70.69 & 51.56 \\
& PRISM  & 0.00 & 5.90 & 70.97 & 50.94 \\
\midrule
\multirow{4}{*}{MedVLThinker-3B}
& raw    & 0.00 & 1.06 & 65.14 & 46.58 \\
& ocr    & 0.00 & 0.96 & 62.14 & 45.11 \\
& g-skip & 0.00 & 4.90 & 59.40 & 42.08 \\
& PRISM  & 0.00 & 4.87 & 58.50 & 41.27 \\
\bottomrule
\end{tabular}}
\end{subtable}\hfill
\begin{subtable}[t]{0.49\textwidth}
\centering
\caption{\(L1\)}
\resizebox{\linewidth}{!}{%
\begin{tabular}{llcccc}
\toprule
Model & Cond. & EM & F1 & Llama-J & Mistral-J \\
\midrule
\multirow{4}{*}{HuluMed-4B}
& raw    & 27.13 & 29.43 & 50.54 & 41.22 \\
& ocr    & 26.95 & 29.21 & 49.67 & 40.07 \\
& g-skip & 28.04 & 30.89 & 55.71 & 46.36 \\
& PRISM  & 27.86 & 30.68 & 55.44 & 46.08 \\
\midrule
\multirow{4}{*}{LLaVA-Med-7B}
& raw    & 0.00 & 6.22 & 54.74 & 39.96 \\
& ocr    & 0.00 & 6.25 & 54.76 & 40.25 \\
& g-skip & 0.00 & 10.24 & 58.82 & 44.26 \\
& PRISM  & 0.00 & 10.29 & 58.67 & 44.08 \\
\midrule
\multirow{4}{*}{MedGemma-1.5-4B}
& raw    & 0.00 & 2.09 & 68.18 & 48.75 \\
& ocr    & 0.00 & 2.12 & 66.37 & 47.69 \\
& g-skip & 0.00 & 5.99 & 68.80 & 47.45 \\
& PRISM  & 0.00 & 5.95 & 69.00 & 46.87 \\
\midrule
\multirow{4}{*}{MedVLThinker-3B}
& raw    & 0.00 & 0.98 & 62.91 & 43.48 \\
& ocr    & 0.00 & 0.94 & 60.44 & 42.94 \\
& g-skip & 0.00 & 4.89 & 58.46 & 41.19 \\
& PRISM  & 0.00 & 4.87 & 57.25 & 40.59 \\
\bottomrule
\end{tabular}}
\end{subtable}

\vspace{0.6em}

\begin{subtable}[t]{0.49\textwidth}
\centering
\caption{\(L2\)}
\resizebox{\linewidth}{!}{%
\begin{tabular}{llcccc}
\toprule
Model & Cond. & EM & F1 & Llama-J & Mistral-J \\
\midrule
\multirow{4}{*}{HuluMed-4B}
& raw    & 26.45 & 28.92 & 53.15 & 40.14 \\
& ocr    & 26.17 & 28.51 & 51.90 & 39.38 \\
& g-skip & 30.12 & 32.45 & 59.73 & 45.15 \\
& PRISM  & 30.27 & 32.57 & 59.44 & 45.17 \\
\midrule
\multirow{4}{*}{LLaVA-Med-7B}
& raw    & 0.00 & 6.51 & 57.76 & 41.02 \\
& ocr    & 0.00 & 6.50 & 57.67 & 41.11 \\
& g-skip & 0.00 & 10.50 & 61.65 & 45.55 \\
& PRISM  & 0.00 & 10.51 & 62.63 & 45.40 \\
\midrule
\multirow{4}{*}{MedGemma-1.5-4B}
& raw    & 0.00 & 2.14 & 68.50 & 49.23 \\
& ocr    & 0.00 & 2.27 & 67.66 & 48.35 \\
& g-skip & 0.00 & 6.14 & 70.32 & 48.28 \\
& PRISM  & 0.00 & 6.09 & 70.18 & 47.72 \\
\midrule
\multirow{4}{*}{MedVLThinker-3B}
& raw    & 0.00 & 1.06 & 66.64 & 43.58 \\
& ocr    & 0.00 & 1.06 & 63.54 & 43.02 \\
& g-skip & 0.00 & 5.01 & 62.28 & 40.04 \\
& PRISM  & 0.00 & 5.01 & 62.47 & 39.91 \\
\bottomrule
\end{tabular}}
\end{subtable}\hfill
\begin{subtable}[t]{0.49\textwidth}
\centering
\caption{\(L3\)}
\resizebox{\linewidth}{!}{%
\begin{tabular}{llcccc}
\toprule
Model & Cond. & EM & F1 & Llama-J & Mistral-J \\
\midrule
\multirow{4}{*}{HuluMed-4B}
& raw    & 15.62 & 19.81 & 55.61 & 47.78 \\
& ocr    & 15.41 & 19.47 & 54.93 & 47.43 \\
& g-skip & 24.62 & 27.91 & 59.55 & 52.86 \\
& PRISM  & 24.45 & 27.66 & 59.92 & 52.63 \\
\midrule
\multirow{4}{*}{LLaVA-Med-7B}
& raw    & 0.00 & 6.44 & 57.14 & 49.41 \\
& ocr    & 0.00 & 6.50 & 57.15 & 49.61 \\
& g-skip & 0.00 & 10.44 & 61.17 & 53.05 \\
& PRISM  & 0.00 & 10.41 & 61.47 & 52.67 \\
\midrule
\multirow{4}{*}{MedGemma-1.5-4B}
& raw    & 0.00 & 1.97 & 69.78 & 60.07 \\
& ocr    & 0.00 & 2.07 & 68.16 & 59.02 \\
& g-skip & 0.00 & 6.02 & 70.63 & 60.47 \\
& PRISM  & 0.00 & 5.95 & 69.91 & 60.15 \\
\midrule
\multirow{4}{*}{MedVLThinker-3B}
& raw    & 0.02 & 1.68 & 67.44 & 53.26 \\
& ocr    & 0.12 & 1.88 & 68.00 & 54.28 \\
& g-skip & 0.00 & 5.46 & 67.67 & 53.26 \\
& PRISM  & 0.00 & 5.49 & 67.14 & 52.77 \\
\bottomrule
\end{tabular}}
\end{subtable}

\vspace{0.35em}
\begin{minipage}{\textwidth}
\footnotesize
\textit{Abbreviations:} ocr = OCR-guided blackout masking; g-skip = skip-connected restoration baseline; PRISM = ClinX-PRISM, the proposed no-skip restoration pathway with mask-guided post-processing.
\end{minipage}
\end{table*}

\subsection{Expanded Model-Wise Results}
\label{app:modelwise_full}

In Table~\ref{tab:appendix_modelwise_levels}, we report the full model-wise utility breakdown across all question-side privacy levels \(L0\) through \(L3\) and all the image conditions. Each sub-table corresponds to one text privacy level and reports exact match (EM), token-level F1, and judge-based utility for each evaluated medical VLM under the four image-side conditions. These results complement the aggregate privacy-utility analysis in the main text by showing whether the observed restoration trends hold consistently across model families and privacy levels, or not.

\

\section{Benchmark Construction and Sanitization Details}
\label{app:benchmark_details}

\subsection{Synthetic PHI Injection and Shared Manifest Construction}
\label{app:phi_injection}

In order to construct a controlled multimodal privacy-stress setting, we first injected the synthetic PHI into both of the image and question modalities. For each sample, we generated a shared manifest \(z_i\) which contained the synthetic identity fields, that includes patient name, date of birth, medical record number, hospital identifier, and image placement metadata. Given an original image \(x_i\), synthetic PHI fields \(z_i\), and a placement region \(r_i\), we then construct the PHI-overlaid image as
\[
\tilde{x}_i = \mathrm{Overlay}(x_i,z_i,r_i),
\qquad
m_i(u,v)=\mathbbm{1}[(u,v)\in r_i].
\]
The same manifest was used to render the image-side PHI overlay, construct the question-side PHI prefix, match OCR detections against injected identifiers, and compute leakage metrics after the sanitization.
\begin{figure}[t]
    \centering
    \begin{subfigure}[t]{0.31\linewidth}
        \centering
        \includegraphics[width=\linewidth]{images/Appendix/ClinX/clean.png}
        \caption{Clean image \(x_i\)}
    \end{subfigure}
    \hfill
    \begin{subfigure}[t]{0.31\linewidth}
        \centering
        \includegraphics[width=\linewidth]{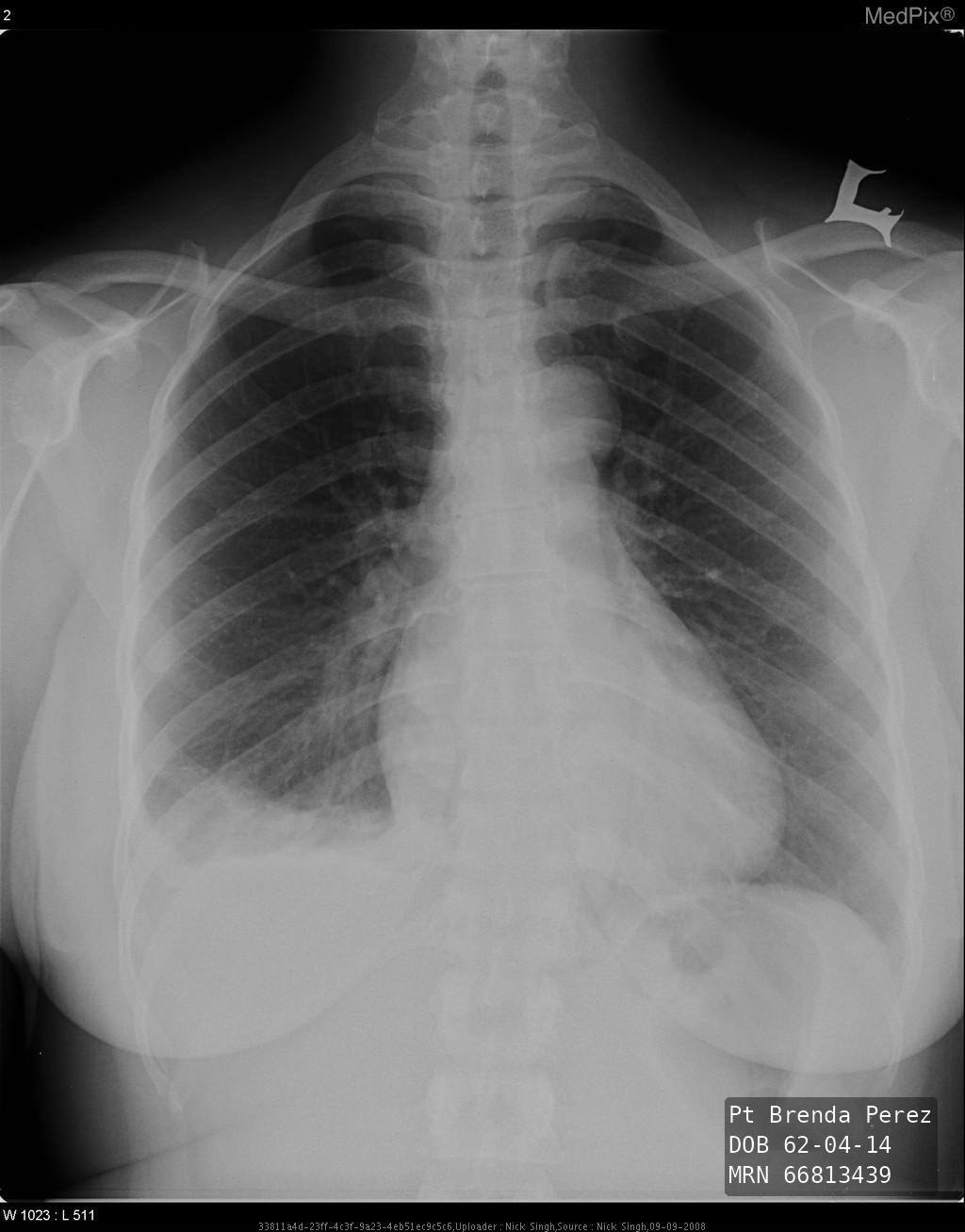}
        \caption{PHI-overlaid image \(\tilde{x}_i\)}
    \end{subfigure}
    \hfill
    \begin{subfigure}[t]{0.31\linewidth}
        \centering
        \includegraphics[width=\linewidth]{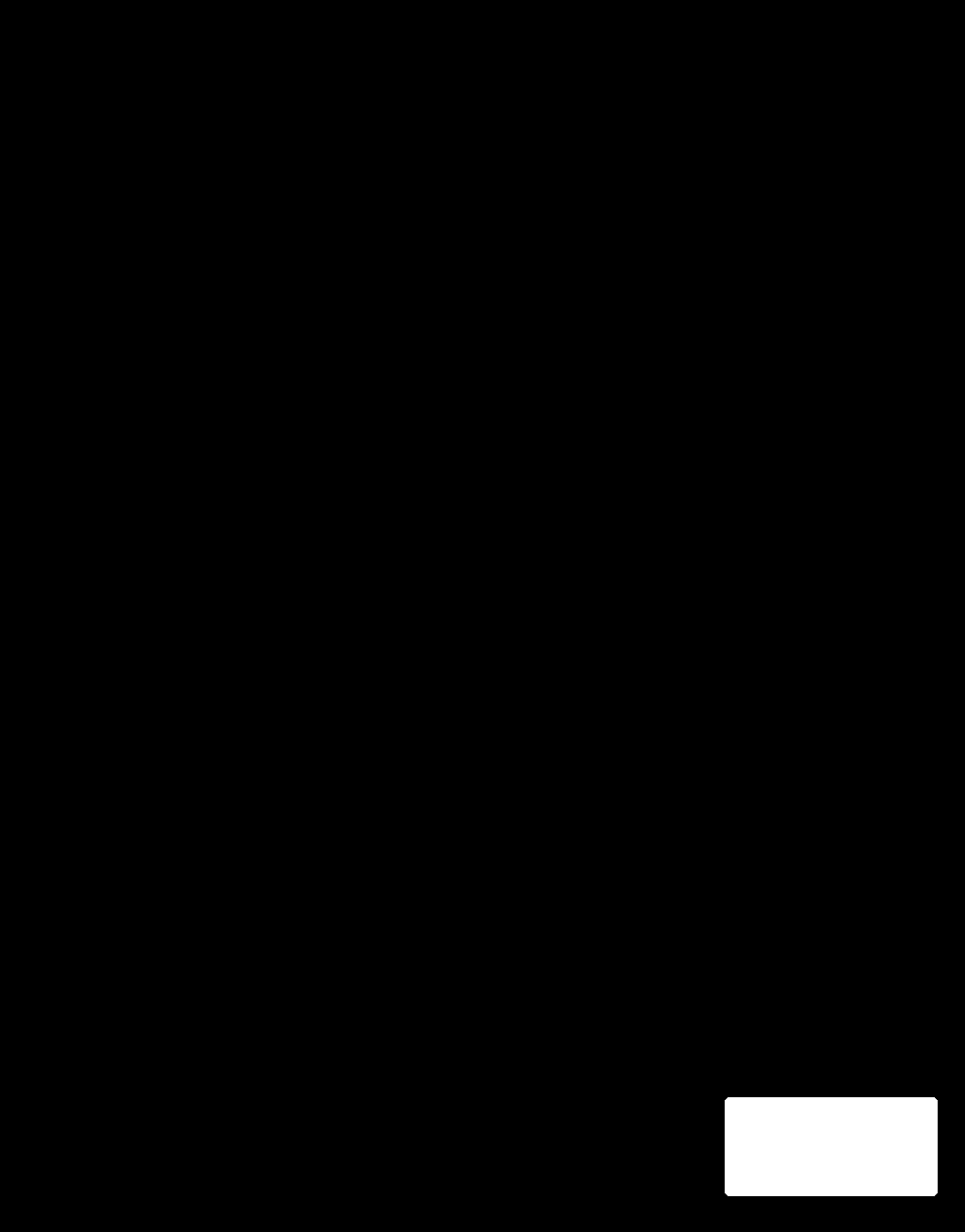}
        \caption{Binary PHI mask \(m_i\)}
    \end{subfigure}

    \caption{Synthetic PHI construction from clean image to PHI-overlaid image and binary mask.}
    \label{fig:phi_synthesis_mask_example}
\end{figure}
Figure~\ref{fig:phi_synthesis_mask_example} presents the image-side construction. Starting from the clean image \(x_i\), synthetic identifiers from the shared manifest are rendered onto a predefined placement region to produce the PHI-overlaid image \(\tilde{x}_i\). The binary mask \(m_i\) identifies the localized sensitive region used by the image-side sanitization branch. As all of the identifiers are synthetic and manifest-defined, the benchmark supports direct measurement of residual PHI leakage without exposing real patient information.

\subsection{OCR Matching and PHI-Only Mask Construction}
\label{app:mask_construction}

For each of the PHI-overlaid image \(\tilde{x}_i\), we first used OCR to produced candidate detections denoted by the \(O_i=\{(s_{ij},b_{ij},c_{ij})\}_{j=1}^{M_i}\), 
where \(s_{ij}\) is the recognized string, \(b_{ij}\) is the bounding region, and \(c_{ij}\) is the OCR confidence. Each of the detected string was first normalized and then we compare them against the synthetic fields in the shared manifest \(z_i\). A detection was treated as PHI if it matched or partially overlapped with a manifest identifier and satisfied the OCR confidence threshold.
The binary PHI-only mask \(m_i\) was then constructed by projecting matched bounding regions into image space and then we applied a small dilation to cover boundary pixels around the rendered text, to ensure that the PHI is fully covered, and the bounding box is seamless. This strategy differs from generic text masking, as only OCR detections corresponding to manifest-defined identifiers were masked, which ensured that only the PHI was masked and not other supporting information of the image itself. If the OCR completely missed a PHI token, the token was not included in the mask; therefore, ClinX does not assume perfect localization. Residual leakage was measured after sanitization using OCR-based privacy metrics.

\subsection{Progressive Question De-Identification Details}
\label{app:text_details}

On the Question-side, the de-identification followed four privacy levels, where \(L0\) retained the PHI-augmented question and served as the weakest text-side privacy condition. \(L1\) applied regex-based masking to explicit surface-form identifiers, including names, dates, medical record numbers, phone numbers, email addresses, and hospital identifiers. \(L2\) extended this process with context-aware masking, targeting identifier-like spans near anchors such as ``Patient,'' ``Name,'' ``DOB,'' ``MRN,'' ``ID,'' and ``Hospital.''
And finally at \(L3\) level of text deidentification, we applied rewrite-based de-identification offline using the Qwen2.5-7B-Instruct~\citep{qwen2.5}. The rewrite prompt instructed the model to remove any identity-bearing information if present at all while also preserving the medical question being asked. In practice, the rewrite was considered valid when the clinical finding, anatomy, imaging modality, disease cue, and question intent were preserved while demographic, institutional, or record-specific identifiers were removed, else we used the original question as the replacement, if the LLMs failed to preserve the core structure of the medical question being asked in level L3. Rewrites were generated just before the downstream VQA evaluation and were not adapted using the predictions of the evaluated VLMs.

\section{Implementation Details}
\label{app:impl}
\subsection{ClinX-PRISM Architecture and Optimization}
\label{app:prism_details}

In ClinX, the PRISM module uses a no-skip, mask-conditioned encoder-decoder generator for the image-side PHI restoration. The PHI-overlaid image, denoted by \(\tilde{x}_i\) is then encoded through a convolutional downsampling path, while the PHI mask \(m_i\) is used as a spatial conditioning signal for the SPADE residual blocks in the bottleneck and decoder. Unlike the skip-connected restoration baseline, ClinX-PRISM does not concatenate encoder features into the decoder. This design forces reconstruction of the PHI region through the mask-conditioned bottleneck rather than through direct low-level feature reuse from the PHI-contaminated input, ensuring that the output does not contain any residual information from the masked region.

\begin{figure*}[t]
    \centering
    \includegraphics[width=0.92\linewidth]{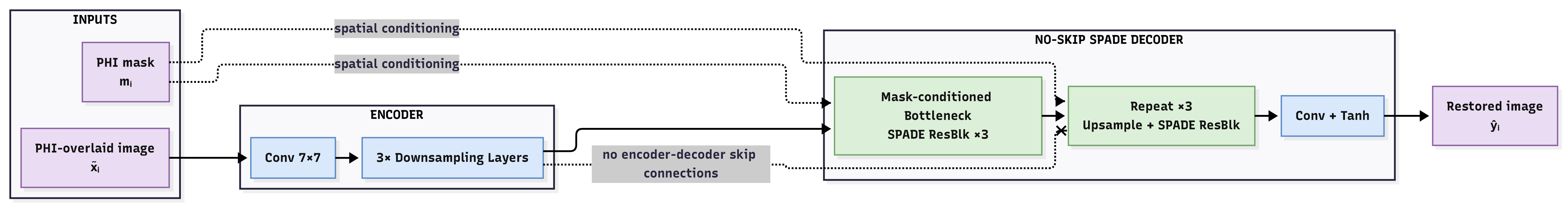}
    \caption{ClinX-PRISM no-skip generator. The PHI mask \(m_i\) conditions the SPADE bottleneck and decoder, while skip connections are removed to force restoration through the mask-conditioned bottleneck.}
    \label{fig:prism_generator}
\end{figure*}
The generated output, denoted by \(\hat{y}_i\) is not used for replacing the complete image. It is only used for generating the pixels within the PHI mask; outside of the mask, the original input or pixels from the PHI-overlaid image is retained. Next, the primary \texttt{PRISM} condition executes the post-processing stage within the mask as discussed in the main body of the paper. The secondary \texttt{PRISM-flat} diagnosis is a variation where no post-processing stage takes place.
For the case of adversarial supervision, we use the PatchGAN architecture for the discriminator network. In contrast to the standard approach of providing a global binary decision with real/fake score, the discriminator generates a spatial realism score map for the input image. This patch-level objective encourages locally plausible restoration in the PHI region while keeping the training objective lightweight and consistent across restoration variants.
\begin{figure*}[t]
    \centering
    \includegraphics[width=\linewidth]{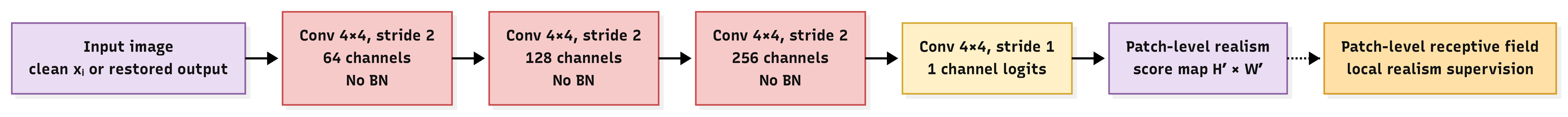}
    \caption{PatchGAN discriminator producing a patch-level realism map for adversarial supervision.}
    \label{fig:patchgan_discriminator}
\end{figure*}
Both of the models (skip-connected baseline and ClinX-PRISM) have been trained using the same adversarial training setting for isolating the effect of information flow through the generator. Tuples of data \((\tilde{x}_i,m_i,x_i)\) were used during the training process, where \(x_i\) refers to the original image before synthetic PHI overlay. The original clean image was used for restoration supervision only during the training process, and no access was made to the original images during evaluation. The objective combined adversarial supervision, masked reconstruction inside the PHI region, identity preservation outside the PHI mask, and total-variation regularization.

\begin{table}[H]
\centering
\caption{ClinX-PRISM training and post-processing configuration.}
\label{tab:prism_config}
\begin{tabular}{ll}
\toprule
Setting & Value \\
\midrule
Input resolution & \(224\times224\) \\
Epochs & 10 \\
Batch size & 6 \\
Learning rate & \(10^{-4}\) \\
Precision & BF16 \\
Hardware & \(2\times\) NVIDIA A100 80GB \\
Discriminator & PatchGAN-style discriminator \\
Generator variants & \texttt{g-skip}, \texttt{PRISM} \\
\midrule
Mask feathering & 10 px Gaussian blur \\
Luminance grain strength & 0.12 \\
Color jitter std. & 0.04 \\
Micro-blur radius & 0.7 px \\
JPEG quality & 92 \\
Edge band & 8 px, reserved \\
\bottomrule
\end{tabular}
\end{table}

The post-processing parameters in Table~\ref{tab:prism_config} are applied only to the \texttt{PRISM} condition. The \texttt{g-skip} baseline uses masked compositing without mask-guided post-processing, the \texttt{ocr} condition applies direct blackout masking, and \texttt{raw} remains unchanged. For reproducibility, the stochastic grain and color-jitter perturbations should be generated with a fixed sample-level seed during final evaluation.

\subsection{LLM-as-a-Judge Prompting and Scoring Protocol}
\label{app:judge_details}

LLM-as-a-Judge scoring was applied as an additional form of evaluation besides exact match and token-level F1 scoring. Each LLM-as-a-Judge was presented with the query, the ground truth short answer, and model prediction, after which it returned either a positive or negative binary score based on whether the predicted answer was clinically consistent with the ground truth. Image, privacy condition, LLM name, and PHIs were not presented to each judge. The prompt was identical for Llama 3.1-8B and Mistral-7B-Instruct-v0.3.
\\
\\
\begin{lstlisting}[
basicstyle=\ttfamily\scriptsize,
breaklines=true,
frame=single,
caption={LLM-as-a-Judge prompt template used for binary clinical equivalence scoring.},
label={lst:judge_prompt}
]
System:
You are a strict medical VQA evaluator. Return only the requested score and no explanation.

User:
Given the following medical visual question answering triple:

Q (Question): <QUESTION>
GT (Ground Truth): <GROUND_TRUTH>
A (Model answer): <MODEL_ANSWER>

Goal:
- Decide if A is clinically and semantically consistent with GT.
- Ignore minor wording or stylistic differences.
- If A is mostly correct and captures the clinical meaning of GT, score 1.
- If A is incorrect, contradicts GT, or misses the main point, score 0.

Important:
- Do NOT be overly lenient.
- Focus on medical correctness and whether the key concept in GT is correctly conveyed.

Respond with exactly one character and no extra text:
1 = clinically and semantically consistent with GT
0 = incorrect, contradictory, or misses the main point
\end{lstlisting}
The final judge score was computed as the percentage of examples assigned score \(1\). We report scores from both judges separately rather than averaging them, since the two models may differ in strictness, due to their architectural design.

\section{Qualitative Examples and Limitations}
\label{app:qual_limit}

\subsection{Qualitative Examples}
\label{app:qual_examples}

\begin{figure*}[t]
    \centering
    \includegraphics[width=0.90\linewidth]{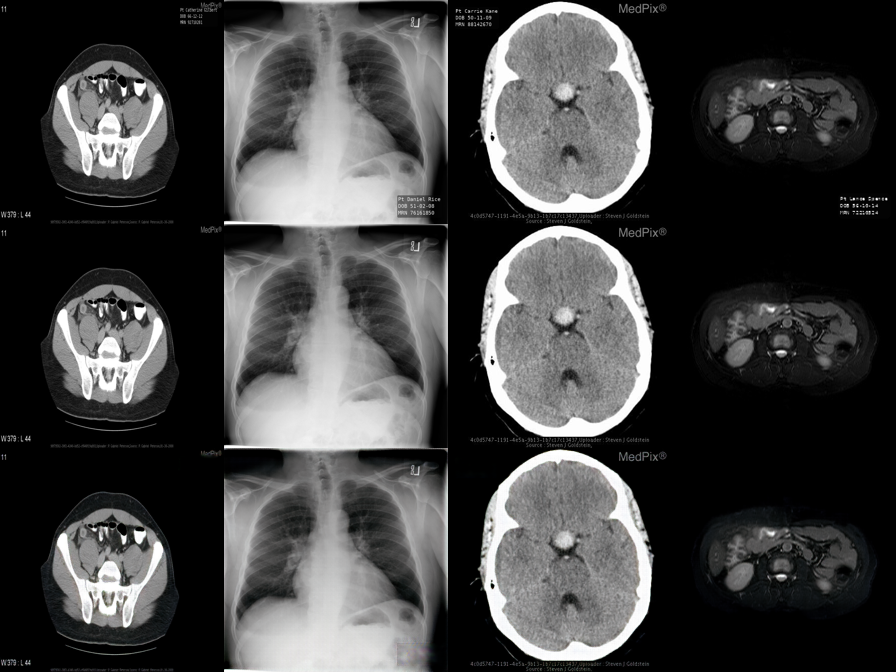}
    \caption{Qualitative ClinX-PRISM restoration examples. Each column shows one sample. Rows show the PHI-overlaid input, clean target used only for paired restoration training, and ClinX-PRISM restored output.}
    \label{fig:qual_image_sanitization}
\end{figure*}

In this section, we present the examples of ClinX-PRISM restoration across different kinds of medical images are demonstrated qualitatively in Fig.~\ref{fig:qual_image_sanitization}. Each of these columns represent one sample, where the first row depicts an input image containing PHI overlay, followed by the clean target image used solely for paired restoration learning, and finally the ClinX-PRISM reconstructed image is shown in the last row. This section demonstrates how ClinX-PRISM learns to restore an image, by reconstructing the image after removing the PHI-corrupted sections without altering other parts of the image. The clean target is shown only to explain the paired training setup and is not available during evaluation.

\subsection{Limitations of the Privacy Evaluation}
\label{app:limitations}
While ClinX aims to mitigate exposure of PHI in the medical datasets through visible image burned-in text and question-side identifiers, but no explicit and formal cryptographic privacy protection is provided. The image-side masks are contingent on OCR detection, and any missed identifier that stays out of the PHI mask will be left in even after sanitization. Residual OCR leakage after sanitization is thus reported instead of assuming perfect PHI localization.
The proposed benchmark employs synthetic PHI injection to avoid handling and exposure of actual PHI present in the real medical dataset, in order to conduct repeatable evaluation of privacy loss. The advantage of this approach is in providing a controlled framework for evaluation, however it may fail to account for some visual and linguistic variances present in real medical images. Similarly, the leakage measures are contingent on OCR recall, with the focus on token-level recovery rather than more comprehensive re-identification risks from non-textual visual content, linguistic characteristics, institutional style, and other demographic shortcuts.
We evaluated ClinX-PRISM module using an adversarial inpainting backbone and a skip-connected restoration ablation. We did not include diffusion-based inpainting models in the primary comparison. Diffusion models may improve perceptual restoration quality, but they introduce additional prompt design, stochastic sampling, computational cost, and reproducibility considerations. A broader comparison with diffusion-based restoration and OCR-aware refinement objectives remains an important direction for future work.


\end{document}